\documentclass{bmvc2k}
\usepackage{amssymb}
\usepackage{amsmath}
\usepackage{algorithm} 
\usepackage{algpseudocode} 
\usepackage{enumitem}

\title{Plot2Spectra: an Automatic Spectra Extraction Tool}

\addauthor{Weixin Jiang}{weixinjiang2022@u.northwestern.edu}{12}
\addauthor{Eric Schwenker}{e.schwenker89@gmail.com}{23}
\addauthor{Trevor Spreadbury}{trevorspreadbury@gmail.com}{2}
\addauthor{Kai Li}{kai1@fb.com}{4}
\addauthor{Maria K.Y. Chan}{mchan@anl.gov}{2}
\addauthor{Oliver Cossairt}{olivercossairt@gmail.com}{1}

\addinstitution{
 Department of Computer Science\\
 Northwestern University\\
 Illinois, USA
}
\addinstitution{
 Center for Nanoscale Materials\\
 Argonne National Laboratory\\
 Illinois, USA
}
\addinstitution{
 Department of Materials Science and Engineering\\
 Northwestern University\\
 Illinois, USA
}
\addinstitution{
 Facebook Reality Lab\\
 Facebook Inc.\\
 California, USA
}

\runninghead{Student, Prof, Collaborator}{BMVC Author Guidelines}

\begin{document}

\maketitle

\begin{abstract}
Different types of spectroscopies, such as X-ray absorption near edge structure (XANES) and Raman spectroscopy, play a very important role in analyzing the characteristics of different materials. In scientific literature, XANES/Raman data are usually plotted in line graphs which is a visually appropriate way to represent the information when the end-user is a human reader.    However, such graphs are not conducive to direct programmatic analysis due to the lack of automatic tools. In this paper, we develop a plot digitizer, named Plot2Spectra, to extract data points from spectroscopy graph images in an automatic fashion, which makes it possible for large scale data acquisition and analysis. Specifically, the plot digitizer is a two-stage framework. In the first axis alignment stage, we  adopt an anchor-free detector to detect the plot region and then refine the detected bounding boxes with an edge-based constraint to locate the position of two axes. We also apply scene text detector to extract and interpret all tick information below the x-axis. In the second plot data extraction stage, we first employ semantic segmentation to separate pixels belonging to plot lines from the background, and from there, incorporate optical flow constraints to the plot line pixels to assign them to the appropriate line (data instance) they encode. Extensive experiments are conducted to validate the effectiveness of the proposed plot digitizer, which shows that such a tool could help accelerate the discovery and machine learning of materials properties. 
\end{abstract}

\section{Introduction}
\label{sec:intro}

Spectroscopy, primarily in the electromagnetic spectrum, is a fundamental exploratory tool in the fields of physics, chemistry, and astronomy, allowing the composition, physical structure and electronic structure of matter to be investigated at the atomic, molecular and macro scale, and over astronomical distances. Among them, XANES and Raman spectroscopy play a very important role in analyzing the characteristics of materials at the atomic level. For the purpose of understanding the insights behind these measurements, data points are usually displayed in graphical form within scientific journal articles. However, it is not standard for materials researchers to release raw data along with their publications. As a result, other researchers have to use interactive plot data extraction tools to extract data points from the graph image, which makes it difficult for large scale data acquisition and analysis. In particular, high-quality experimental spectroscopy data is critical for the development of machine learning (ML) models, and the difficulty involved in extracting such data from the scientific literature hinders efforts in ML of materials properties. It is therefore highly desirable to develop a tool for the digitization of spectroscopy graphical plots. We use as prototypical examples XANES and Raman spectroscopy graphs, which often have a series of difficult-to-separate line plots within the same image. However, the approach and tool can be applied on other types of graph images. 

WebPlotDigitizer \cite{Rohatgi2020} is one of the most popular plot data extraction tools to date. However, the burden of having to manually align axes, input tick values, pick out the color of the target plot and draw the region where the plot falls in is cumbersome and not conducive to automation.

\begin{figure}[!t]
\setlength{\belowcaptionskip}{-15pt}
\setlength{\abovecaptionskip}{-20pt}
\centering
\includegraphics[width=0.95\columnwidth]{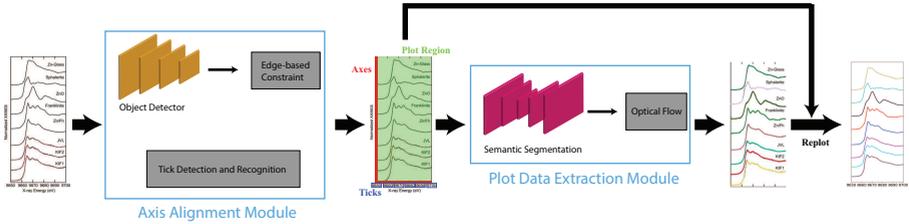}
\caption{An overview of the proposed Plot2Spectra. An example of XANES graph images is first fed into the axis alignment module, which outputs the position of axes, the values of the ticks along x axis and the plot region. Then the plot region is fed into the plot data extraction module which detects plot lines.}
\label{fig: pipeline}
\end{figure}

In this paper, we develop Plot2Spectra, which transforms plot lines from the graph image into sets of coordinates in an automatic fashion. As shown in Fig. \ref{fig: pipeline}, there are two stages in the plot digitizer. The first stage involves an axis alignment module. We first adopt an anchor-free object detection model to detect plot regions, and then refine the detected bounding boxes with the edge-based constraint to force the left/bottom edges to align with axes. Then we apply the scene text detection and recognition algorithms to recognize the ticks along the x axis. The second stage is the plot data extraction module. We first employ semantic segmentation to separate pixels belonging to plot lines from the background, and from there, incorporate optical flow constraints to the plot line pixels to assign them to the appropriate line (data instance) they encode. 

The contribution of this paper is summarized as follows.
\begin{enumerate}[nolistsep]
  \item To the best of our knowledge, we are the first to develop the plot digitizer which extracts data points from the graph image in a fully automatic fashion.
  \item We suppress axis misalignment by introducing an edge-based constraint to refine the bounding boxes detected by the conventional CNN-based object detection model.
  \item We propose an optical flow based method, by analyzing the statistical proprieties of plot lines, to address the problem of plot line detection (i.e. assign foreground pixels to the appropriate plot lines).
\end{enumerate}

\section{Related Work}
\label{sec:related_work}

\subsection{Object detection}

Object detection aims at locating and recognizing objects of interest in the given image, which generates a set of bounding boxes along with their classification labels. CNN-based object detection models can be classified into two categories: anchor-based methods \cite{ren2015faster,redmon2017yolo9000,redmon2018yolov3} and anchor-free methods \cite{redmon2016you,zhou2017east, tian2019fcos,wang2019region,yang2018metaanchor,zhang2021learning}. Anchors are a set of predefined bounding boxes and turn the direct prediction problem into the residual learning problem between the pre-assigned anchors and the ground truth bounding boxes because it is not trivial to directly predict an order-less set of arbitrary cardinals. However, anchors are usually computed by clustering the size and aspect ratio of the bounding boxes in the training set, which is time consuming and likely to fail if the morphology of the object varies dramatically. To address this problem, anchor-free methods either learn custom anchors along with the training process of the detector \cite{zhang2021learning} or reformulate the detection in a per-pixel prediction fashion \cite{tian2019fcos}. In this paper, since the size and aspect ratio of plot regions vary dramatically, we build our axis alignment module with anchor-free detectors.

\subsection{Scene text detection and recognition}

Scene text detection aims at locating the text information in a given image. Early text detectors use box regression adapted from popular object detectors \cite{ren2015faster,long2015fully}. Unlike natural objects, texts are usually presented in irregular shapes with various aspect ratios. Deep Matching Prior
Network (DMPNet) \cite{liu2017deep} first detect text with quadrilateral sliding window and then refine the bounding box with a shared Monte-Carlo method. Rotation-sensitive Regression Detector (RDD) \cite{liao2018rotation} extracts rotation-sensitive features with active rotating filters \cite{zhou2017oriented}, followed by a regression branch which predicts offsets from a default rectangular box to a quadrilateral. More recently, character-level text detectors are proposed to first predict a semantic map of characters and then predict the association between these detected characters. Seglink \cite{shi2017detecting} starts with character segment detection and then predicts links between adjacent segments. Character Region Awareness For Text detector (CRAFT) \cite{baek2019character} predicts the region score for each character along with the affinity score between adjacent characters. 

Scene text recognition aims at recognizing the text information in a given image patch. A general scene text recognition framework usually contains four stages, normalizing the text orientation \cite{jaderberg2015spatial}, extracting features from the text image\cite{he2016deep,simonyan2014very}, capturing the contextual information within a sequence of characters \cite{shi2016end,shi2016robust}, and estimating the output character sequence \cite{cheng2017focusing}. In this paper, we apply a pre-trained scene text detection model \cite{baek2019character} and a pre-trained scene text recognition model \cite{baek2019wrong} to detect and recognize text labels along the x axis, respectively. We focus on the x-axis labels because in spectroscopy data, the y-axis labels are often arbitrary, and only relative intensities are important. However, the general framework presented can be extended in the future to include y-axis labels. 

\subsection{Instance segmentation}

The goal of instance segmentation is to assign different semantic labels to each pixel in the given image and group pixels into different instances. Instance segmentation algorithms can usually be divided into two categories: proposal-based methods and proposal-free methods. Proposal-based methods \cite{he2017mask, liu2018path} address the instance segmentation by first predicting object proposals (i.e. bounding boxes) for each individual instance and then assigning labels to each pixel inside the proposals (i.e. semantic segmentation). The success of the proposal-based methods relies on the morphology of the target object, and is likely to fail if the object is acentric or if there is significant overlap between different instances. Proposal-free methods \cite{neven2018towards,de2017semantic,kong2018recurrent,newell2016associative,neven2019instance} first take segmentation networks to assign different semantic labels to each pixel, then map pixels in the image to feature vectors in a high dimensional embedding space. In this embedding space, feature vectors corresponding to pixels belonging to the same object instance are forced to be similar to each other, and feature vectors corresponding to pixels belonging to different object instances are forced to be sufficiently dissimilar.  However, it is difficult to find such an embedding space if the objects do not have rich features, such as the plot lines in graph images. In this paper, we customize our plot data extraction module by replacing the pixel embedding process with an optical flow based method, which groups data points into plot lines with continuity and smoothness constraints.

\begin{figure}[!t]
\setlength{\belowcaptionskip}{-15pt}
\setlength{\abovecaptionskip}{-20pt}
\centering
\includegraphics[width=0.95\columnwidth]{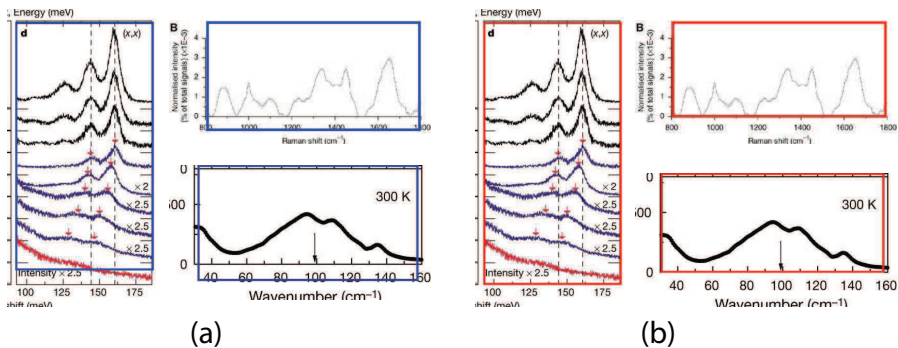}
\caption{Examples of axis misalignment. (a) There are noticeable gaps between the left/bottom edges of the bounding boxes (blue boxes) and theaxes. (b) The left/bottom edges of the bounding boxes (red boxes) are perfectly aligned with the the axes.}
\label{fig: misalignment}
\end{figure}

\section{Method}
\label{sec:method}

In this section, we provide more details about the proposed Plot2Spectra tool. The general pipeline of the proposed method is shown in Fig. \ref{fig: pipeline}. The pipeline is made of two modules. The first module is the axis alignment module, which takes the graph image as the input and outputs the position of axes, the value and position of ticks along x axis as well as a suggested plot region. The second module is the plot data extraction module, which takes the plot region as the input and outputs each detected plot line as a set of (x, y) coordinates. With the detected plot lines, ticks, and axes, we are able to perform any subsequent plot analysis (e.g. re-plot data into a new graph image, compute similarities, perform ML tasks, etc.).

\subsection{Axis alignment}

In the axis alignment module, we first adopt an anchor-free object detector \cite{tian2019fcos} to detect plot regions and refine the predicted bounding boxes with the edge-based loss. We then apply the pre-trained scene text detector \cite{baek2019character} and the pre-trained scene text recognizer \cite{baek2019wrong} to extract and interpret all tick labels below the x-axis.

Given the graph image $I \in \mathbb{R}^{H\times W\times 3}$, let $F \in \mathbb{R}^{H_F\times W_F\times C}$ be the feature map computed by the backbone CNN. Assume the ground truth bounding boxes for the graph image are defined as $\{B_i\}$, where $B_i = (x_0^i, y_0^i, x_1^i, y_1^i) \in \mathbb{R}^4$. Here $(x_0^i, y_0^i)$ and $(x_1^i, y_1^i)$ denote the coordinates of the left-top and right-bottom corners of the bounding box, respectively. For each location $(x, y)$ on the feature map $F$, it can be mapped back to the graph image as $(\frac{W}{W_F}(x+\frac{1}{2}), \frac{H}{H_F}(y+\frac{1}{2}))$ (i.e. the center of the receptive field of the location $(x, y)$). For the feature vector at each location $(x, y)$, a 4D vector $t_{x,y} = (l, t, r, b)$ and a class label $c_{x,y}$ are predicted. The ground truth class label is denoted as $c_{x,y}^\ast = \{0, 1\}$ (i.e. $0, 1$  denote the labels for background and foreground pixels, respectively) and the ground truth regression targets for each location is denoted as $t_{x,y}^\ast = \{ l^\ast, t^\ast, r^\ast, b^\ast\ | l^\ast = x-x_0^i, t^\ast = y-y_0^i, r^\ast = x_1^i-x, b^\ast = y_1^i-y \}$. Then the loss function for the detector comprises a classification loss and a bounding box regression loss
\begin{equation}
\mathcal{L}^{det} = \frac{1}{N_{pos}}\sum_{x,y}\mathcal{L}_{cls}(c_{x,y}, c_{x,y}^\ast) + \mathbf{1}_{\{c_{x,y}^\ast>0\}}\mathcal{L}_{reg}(t_{x,y}, t_{x,y}^\ast)
\label{eq: loss_det}
\end{equation}
where $\mathcal{L}_{cls}$ denotes the focal loss in \cite{lin2017focal} and $\mathcal{L}_{reg}$ denotes the IoU (Intersection over Union) loss in \cite{yu2016unitbox}. $N_{pos}$ denotes the number of locations that fall into any ground truth box. $\mathbf{1}_{\{\cdot\}}$ is the indicator function, being 1 if the condition is satisfied and 0 otherwise.

However, the left and bottom edges of the predicted bounding boxes by the detector may not align with the axes, as shown in Fig. \ref{fig: misalignment}(a). Therefore, we introduce an edge-based constraint to force the left/bottom edges of the detected bounding boxes to align with axes inspired by the observation that the values of pixels along axes usually stay constant.
\begin{equation}
\mathcal{L}^{edge}(x,y) = \sum_{u=x-l}^{x+r}I(u, y+b) + \sum_{v=y-t}^{y+b}I(x-l, v), \quad s.t. \; (l, t, r, b) \in t_{x,y}
\label{eq: L_edge}
\end{equation}
The first term forces the left edge to have constant values and the second term forces the bottom edge to have constant values. Then, the axis alignment module is optimized with both the detection loss and the edge-based loss:
\begin{equation}
\mathcal{L}^{AL} = \mathcal{L}^{det} + \mathcal{L}^{edge}
\label{eq: L_axis_alignment}
\end{equation}

However, the edge-based loss term is not differentiable, which means the Eq. \ref{eq: L_axis_alignment} cannot be optimized directly. In practice, we take the one-step Majorization-Minimization strategy to solve the problem.
\begin{equation}
\underset{l,b}{\arg\min}\{\mathcal{L}^{edge} | \underset{t_{x,y}}{\arg\min}\mathcal{L}^{det}\}
\label{eq: L_MM_solver}
\end{equation}

As shown in Eq. \ref{eq: L_MM_solver}, we first optimize the detection model with the gradient descent method to generate bounding boxes with high confidence scores, then we refine the left/bottom edges of the detected bounding boxes via a Nearest Neighbor search. In particular, we apply the probabilistic Hough transform \cite{kiryati1991probabilistic} to detect lines (i.e. axes candidates) in the graph image and then search for the most confident candidates. Intuitively, the best candidates should be either horizontal or vertical, long enough and close to the edges of the detected bounding box.
\begin{equation}
\begin{split}
L^\ast &= \underset{L_i \in \mathcal{H}(I)}{\arg\min} D_{dist}(L_i, E) \\
s.t. \quad \|D_{angle}(L_i, &E)\|_2^2 > \epsilon_1, \;  D_{length}(L_i, E) > \epsilon_2
\end{split}
\label{eq: hough_transform}
\end{equation}
where $\mathcal{H}$ denotes the probabilistic Hough transform operator, $E \in \{E^{left}, E^{bottom}\}$ denotes the left or bottom edge of the bounding box. $D_{angle}$ measures the cosine similarity between the given two lines. $D_{length}$ measures the ratio between the length of the detected line and the edge, and $D_{dist}$ measures the horizontal/vertical distance between the two parallel lines. Empirically,  $\epsilon_1$ and $\epsilon_2$ are set to be 0.98 and 0.5, respectively.


\begin{figure}[!t]
\setlength{\belowcaptionskip}{-15pt}
\setlength{\abovecaptionskip}{-20pt}
\centering
\includegraphics[width=0.95\columnwidth]{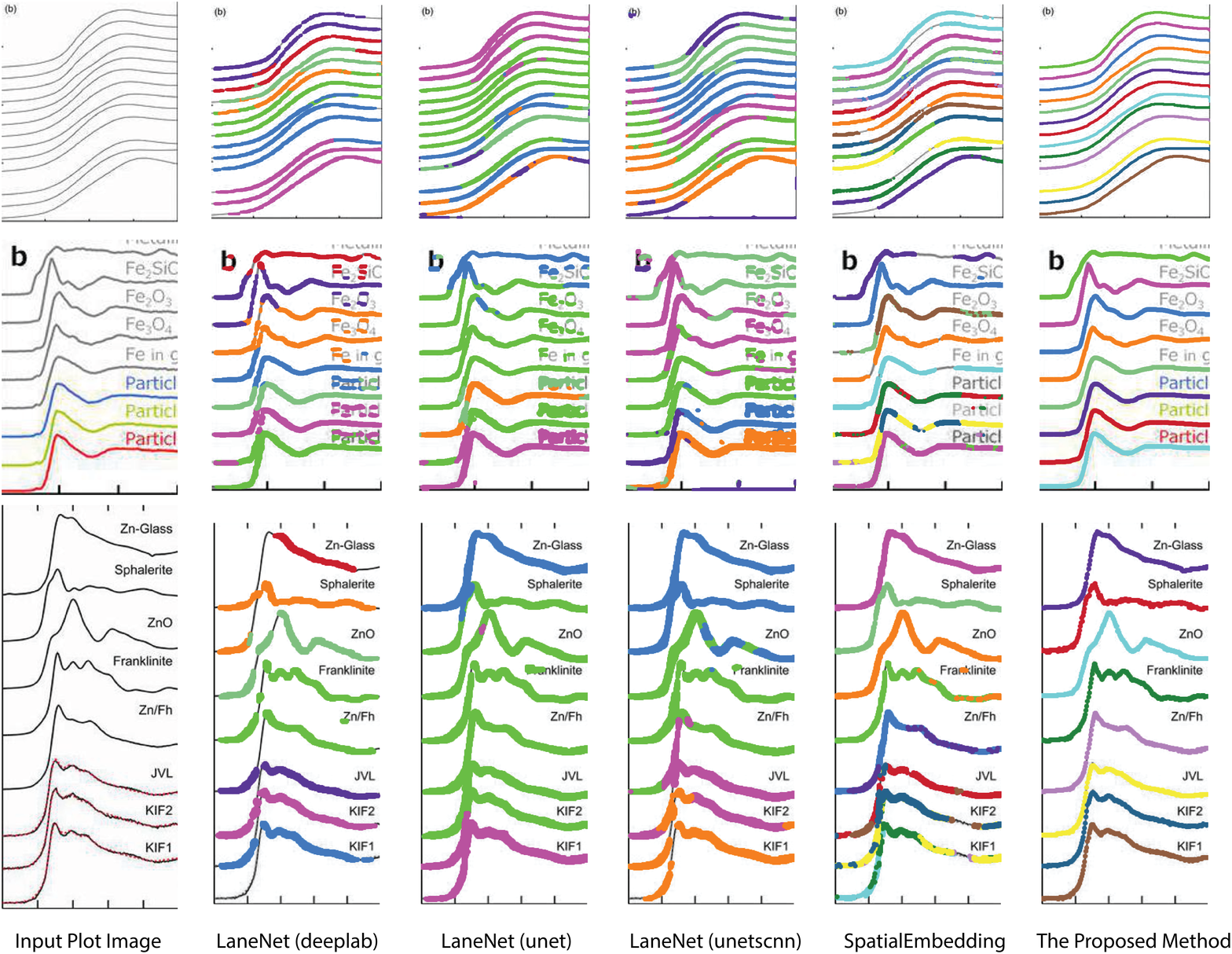}
\caption{Comparison of the results of plot data extraction between instance segmentation algorithms and the proposed method. The first column is the input plot images and the rest columns are plot line detection results with different methods. All pixels belonging to a single line (data instance) should be assigned to the same color. }
\label{fig: visual_comparison}
\end{figure}
\subsection{Plot data extraction}

In the plot data extraction module, we first perform semantic segmentation to separate pixels belonging to plot lines from the background, and from there, apply optical flow constraints to the plot line pixels to assign them to the appropriate line (data instance) they encode. 

\begin{equation}
\mathcal{L}^{seg} = -C\log(\overset{\sim}{C})-(1-C)(1-\log(\overset{\sim}{C}))
\label{eq: binary_segmentation}
\end{equation}
where $\overset{\sim}{C} = S(I^p) \in \mathbb{R}^{H^p\times W^p}$ denotes the probability map, which is computed by the semantic segmentation model $S$ from the given plot image $I^p$. $C = \{c_i\} \in \mathbb{R}^{H^p\times W^p}$ denotes the ground truth semantic map, $c_i$ is 1 if it is a foreground pixel and otherwise 0.

Pixel embedding in conventional instance segmentation is likely to fail because the plot lines often do not have a sufficient number of distinguishable features between different instances. As shown in Fig. \ref{fig: visual_comparison}, all pixels belonging to a single line (data instance) should be assigned to the same color.  One common mode of failure in conventional instance segmentation models (i.e. LaneNet \cite{neven2018towards}, SpatialEmbedding \cite{neven2019instance}) involves assigning multiple colors to pixels within a single line. Another common failure mode is a result of misclassifications of the pixels during segmentation (see second row in Fig. \ref{fig: visual_comparison}, LaneNet misclassifies background pixels as foreground and SpatialEmbedding misclassifies foreground pixels as background). 

Intuitively, plot lines are made of a set of pixels of the same value and have some statistical properties, such as smoothness and continuity. Here, we formulate the plot line detection problem as tracking the trace of a single point moving towards the y axis direction as the value of x increases. In particular, we introduce an optical flow based method to solve this tracking problem.
\begin{equation}
I^p(x,y) = I^p(x+dx, y+dy) 
\label{eq: optical_flow_assumption}
\end{equation}

Brightness constancy, a basic assumption of optical flow, requires the intensity of the object to remain constant while in motion, as shown in Eq. \ref{eq: optical_flow_assumption}. Based on this property, we introduce the intensity constraint to force the intensity of pixels to be constant along the line.
\begin{equation}
\mathcal{L}^{intensity} = \sum_{i=0}^{W^p-1}\|I^p(x_{i+1}, y_{i+1}) - I^p(x_i, y_i)\|^2_2
\label{eq: L_line_intensity}
\end{equation}

Then we apply a first order Taylor expansion to Eq. \ref{eq: optical_flow_assumption}, which estimates the velocity of the point towards y axis direction at different positions. Based on this, we introduce the smoothness constraint to force the plot line to be differentiable everywhere, i.e.
\begin{equation}
\begin{split}
V(x,y) &= -\frac{I^p_x(x,y)}{I^p_y(x,y)} \\
\mathcal{L}^{smooth} &= \sum_{i=0}^{W^p-1}\|y_{i+1}-y_i-V(x_i,y_i)\|^2_2
\end{split}
\label{eq: L_line_smooth}
\end{equation}
where $I^p_x, I^p_y$ denote the gradient map along x-direction and y-direction, respectively. $V(x,y)$ denotes the velocity of the point along y-direction. 

Also, we introduce the semantic constraint to compensate the optical flow estimation and force more foreground pixels to fall into the plot line.
\begin{equation}
\mathcal{L}^{semantic} = \sum_{i=0}^{W^p}\|1 - \overset{\sim}{C}(x_i, y_i)\|^2_2
\label{eq: L_line_semantic}
\end{equation}

Therefore, the total loss for plot line detection is
\begin{equation}
\mathcal{L}^{line} = \mathcal{L}^{smooth} + \mathcal{L}^{intensity} + \mathcal{L}^{semantic}
\label{eq: L_line_total}
\end{equation}

\begin{figure}[!t]
\setlength{\belowcaptionskip}{-15pt}
\setlength{\abovecaptionskip}{-20pt}
\centering
\includegraphics[width=0.95\columnwidth]{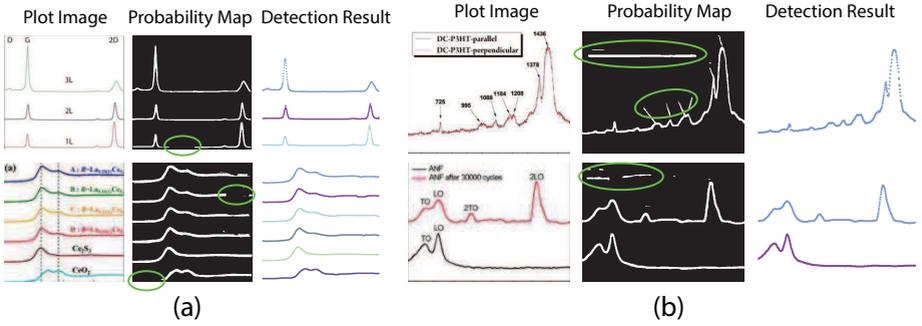}
\caption{Plot line detection with imperfect semantic segmentation results. (a). Part of plot lines is missing in the probability map. (b). Background pixels are misclassified as foreground.}
\label{fig: more_results}
\end{figure}

\begin{figure}[!t]
\setlength{\belowcaptionskip}{-10pt}
\setlength{\abovecaptionskip}{-30pt}
\centering
\includegraphics[width=1.\columnwidth]{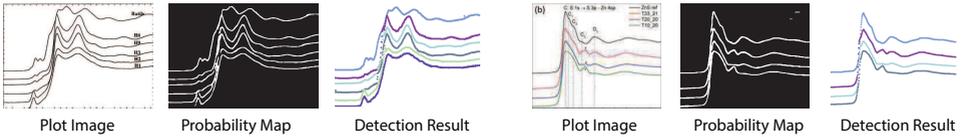}
\caption{Plot line detection with hard examples.}
\label{fig: hard_examples}
\end{figure}

\section{Experiments}
\label{sec: experiments}
In this section, we conduct extensive experiments to validate the effectiveness and superiority of the proposed method.

We collected a large number of graph images from literature using the EXSCLAIM! pipeline \cite{jiang2021two,schwenker2021exsclaim} with the keyword "Raman" and "XANES". Then we randomly selected 1800 images for the axis alignment task, with 800 images for training and 1000 images for validation. For the plot data extraction task, we labeled 236/223 plot images as the training/testing set with LabelMe \cite{labelme2016}.

During the training process, we implement all baseline object detection models \cite{tian2019fcos, wang2019region, zhang2021learning} with the MMDetection codebase \cite{chen2019mmdetection}. The re-implementations strictly follow the default settings of MMDetection. All models are initialized with pre-trained weights on the MS-coco dataset and then fine tuned with SGD optimizer with the labeled dataset for 1000 epochs in total, with initial learning rate as 0.005. Weight decay and momentum are set as 0.0001 and 0.9, respectively. We train the semantic segmentation module from \cite{neven2019instance} in a semi-supervised manner. In particular, we simulate plot images with variations on the number/shape/color/width of plot lines, with/without random noise/blur and then we train the model alternatively with the simulated data and real labeled data for 1000 epochs. Readers may refer to Appendix A in the supplementary material for more implementation details of the optical flow based method.

Visual comparison on plot line detection between the proposed method and conventional instance segmentation algorithms is shown in Fig. \ref{fig: visual_comparison}. In particular, we have 4 baseline methods: LaneNet \cite{neven2018towards} with Deeplab \cite{chen2017deeplab} as the backbone, LaneNet with Unet \cite{ronneberger2015unet} as the backbone, LaneNet with Unetscnn \cite{pan2017spatial} as the backbone and SpatialEmbedding \cite{neven2019instance}. All the instance segmentation algorithms fail to distinguish pixels from different plot lines especially when the number of lines increases and the distance between adjacent lines decreases (e.g. first row in Fig. \ref{fig: visual_comparison}). As expected, the proposed optical flow based method correctly groups pixels into plot lines. Moreover, the proposed method still works even with imperfect semantic segmentation prediction. As shown in Fig. \ref{fig: more_results}, the proposed method is able to inpaint the missing pixels and eliminate misclassified background pixels. The proposed plot line detection method also works for cases containing significant overlap between different plot lines - conditions that can even be challenging for humans to annotate or disambiguate. This is shown in Fig. \ref{fig: hard_examples}.

\begin{figure}[!t]
\setlength{\belowcaptionskip}{-10pt}
\setlength{\abovecaptionskip}{-25pt}
\centering
\includegraphics[width=0.95\columnwidth]{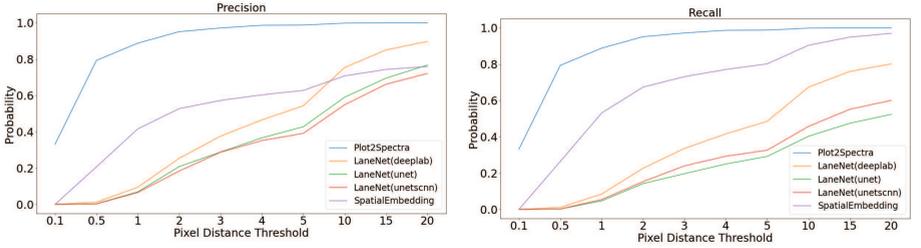}
\caption{Quantitative comparison of plot data extraction between different algorithms.}
\label{fig: quantitative_result_comparison}
\end{figure}

Meanwhile, we introduce a quantitative metric to evaluate the performance of different methods. Here,  $\{L^{pred}\}$ and $\{L^{gt}\}$ denote the set of detected plot lines and ground truth plot lines in the image, respectively. Then we define matched plot lines as $\{L^{match} | L \in \{L^{pred}\}, \underset{\tau \in \{L^{gt}\}}{\min} |L-\tau|  < \epsilon_p \}$ while each ground truth plot line can have at most one matched plot line. Thus we have $\mathrm{Precision} = \frac{\|\{L^{match}\}\|}{\|\{L^{pred}\}\|}, \; \mathrm{Recall} = \frac{\|\{L^{match}\}\|}{\|\{L^{gt}\}\|}$, where $\|.\|$ denotes the number of plot lines in the set and $\epsilon_p$ denotes the threshold of the mean absolute pixel distance between two plot lines. By setting different $\epsilon_p$, we measure the performance of different algorithms, as shown in Fig. \ref{fig: quantitative_result_comparison}. As expected, the proposed algorithm achieves better precision and recall accuracy than the other methods. In particular, there are 935 plot lines in the testing set. Given $\epsilon_p=1$, the proposed plot digitizer detects 831 matched plot lines and given  $\epsilon_p=2$, the proposed plot digitizer detects 890 matched plot lines.




\begin{figure}[!t]
\setlength{\belowcaptionskip}{-15pt}
\setlength{\abovecaptionskip}{-20pt}
\centering
\includegraphics[width=0.95\columnwidth]{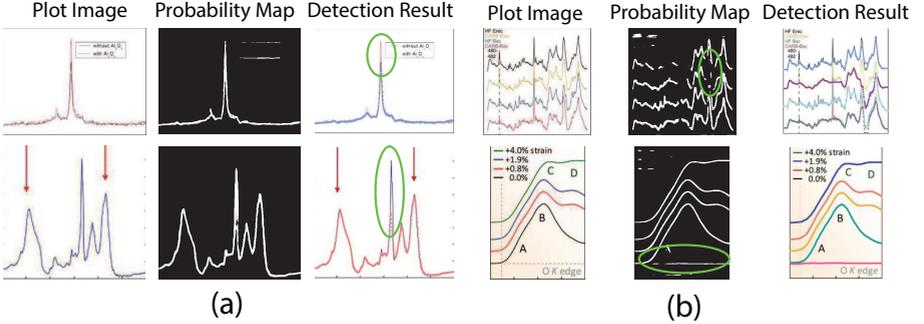}
\caption{Failure cases. (a) Plot2Spectra fails when there is a significant peak. (b) Plot2Spectra fails when a large portion of background/foreground pixels are misclassified.}
\label{fig: failure_cases}
\end{figure}

\section{Conclusion and discussion}
\label{sec:conclusion_and_discussion}

In this paper, we propose the Plot2Spectra to extract data points from XANES/Raman graph spectroscopy images and transform them into coordinates, which enables large scale data collection,  analysis, and machine learning of these types of spectroscopy data. Extensive experiments validate the effectiveness and superiority of the proposed method, even for very challenging examples. Readers may refer to the supplementary material for more ablation studies.

Unfortunately, there are some cases that the proposed model is likely to fail. As shown in Fig. \ref{fig: failure_cases}(a), the current model fails to detect sharp peaks. This is because the first order Taylor approximation in Eq. \ref{eq: L_line_smooth} does not hold for large displacement. A possible way to address this issue is to stretch the plot image along x-direction, which reduces the slope of the peaks. Another kind of failure cases is shown in Fig. \ref{fig: failure_cases}(b). Since the proposed plot line detection algorithm detects plot line in a sequential manner, the error from the previous stage (i.e. semantic segmentation) would affect the performance of the subsequent stage (i.e. optical flow based method). Even though the optical flow based method is robust if some pixels are misclassified (as shown in Fig. \ref{fig: more_results}), significant error in the probability map would still result in a failure. A possible way to address this issue could be training a more advanced semantic segmentation model with more labeled data.

\begin{algorithm}
	\caption{Optical Flow based Algorithm for Plot Line Detection}
	\begin{algorithmic}[]
	    \State \textbf{Initialization:} $V(x,y),\; \overset{\sim}{C},\; I^p(x,y)$
	    \State Pick a point from each plot line as the start position $\{(x_t, y_t^k) | \; k=1,2,...,M\}$, where $M$ is the number of plots
	    \While{$t < W^p$}
	    \State $\hat{y}_{t+1}^k \leftarrow y_t^k + V(x_t, y_t^k)$ \Comment{Estimate the next position of the point with optical flow}
	    
	    \State $y_{cand} \leftarrow \{\overset{\sim}{C}(x_{t+1}) == 1\}$
	    \Comment{Select pixels of plot data in the semantic map}
	    
	    \State $\{\hat{y}_{t+1}^k\} \leftarrow \mathbf{SemanticMap}(\{\hat{y}_{t+1}^k\}, y_{cand})$ 
	    
	    \State $\{y_{t+1}^k\} \leftarrow \mathbf{ColorMap}(x_{t+1}, \{\hat{y}_{t+1}^k\})$
	    
	    \State $t \leftarrow t+1$
	    \EndWhile
	    
	    \While{$t > 0$}
	    \State $\hat{y}_{t+1}^k \leftarrow y_t^k - V(x_t, y_t^k)$ \Comment{Estimate the next position of the point with optical flow}
	    
	    \State $y_{cand} \leftarrow \{\overset{\sim}{C}(x_{t-1}) == 1\}$
	    \Comment{Select pixels of plot data in the semantic map}
	    
	    \State $\{\hat{y}_{t-1}^k\} \leftarrow \mathbf{SemanticMap}(\{\hat{y}_{t-1}^k\}, y_{cand})$ 
	    
	    \State $\{y_{t-1}^k\} \leftarrow \mathbf{ColorMap}(x_{t-1}, \{\hat{y}_{t-1}^k\})$
	    \State $t \leftarrow t-1$
	    \EndWhile
	\end{algorithmic} 
	\label{algorithm: optical_flow_based_propagation}
\end{algorithm}

\section{Appendix A: implementation of the optical flow based algorithm}
\label{sec:appendix_A}

The optical flow based method is implemented as shown in Algorithm. \ref{algorithm: optical_flow_based_propagation}. 
\begin{equation}
\begin{split}
\mathbf{SemanticMap}(\hat{y}^k, y_{cand}) &= \begin{cases} 
    \hat{y}^k, \quad \mathrm{if} \underset{y\in y_{cand}}{\min} \|y-\hat{y}^k\|_2^2 < \Delta_s \\
      \underset{y\in y_{cand}}{\arg\min} \|y-\hat{y}^k\|_2^2, \quad \mathrm{Otherwise}
   \end{cases} \\
\mathbf{ColorMap}(x, \hat{y}^k) &=\begin{cases} 
    \hat{y}^k, \quad \mathrm{if} \underset{y\in \mathcal{N}(\hat{y}^k)}{\min} \|I^p(x,y)-\{I^p(x,\hat{y}^k)\}\|_2^2 < \Delta_c \\
      \underset{y\in \mathcal{N}(\hat{y}^k)}{\arg\min} \|I^p(x,y)-\{I^p(x,\hat{y}^k)\}\|_2^2, \quad \mathrm{Otherwise}
   \end{cases} 
\end{split}
\label{eq: map}
\end{equation}
where $\mathcal{N}(y)$ denotes the neighborhood of $y$, all values in the interval between $y-\delta$ and $y+\delta$. Empirically,  $\delta$ is 10 in this paper. $\Delta_s, \Delta_c$ are two thresholds, which help to suppress imperfection in the probability map (e.g. reject misclassified background pixels and inpaint missing foreground pixels).

\begin{figure}[!t]
\setlength{\belowcaptionskip}{-10pt}
\setlength{\abovecaptionskip}{-10pt}
\centering
\includegraphics[width=0.95\columnwidth]{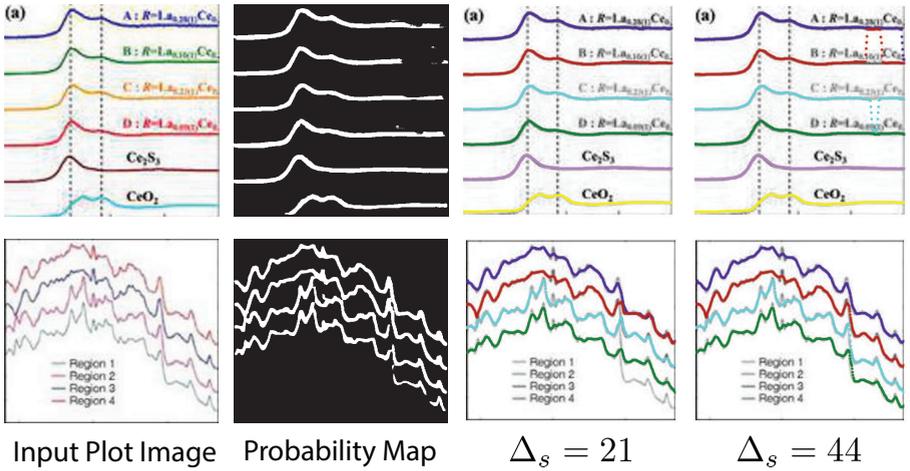}
\caption{Results of plot line detection with different $\Delta_s$. Too large or too small values fail to have a correct detection.}
\label{fig: delta_s}
\end{figure}

\begin{figure}[!t]
\setlength{\belowcaptionskip}{-15pt}
\setlength{\abovecaptionskip}{-15pt}
\centering
\includegraphics[width=0.95\columnwidth]{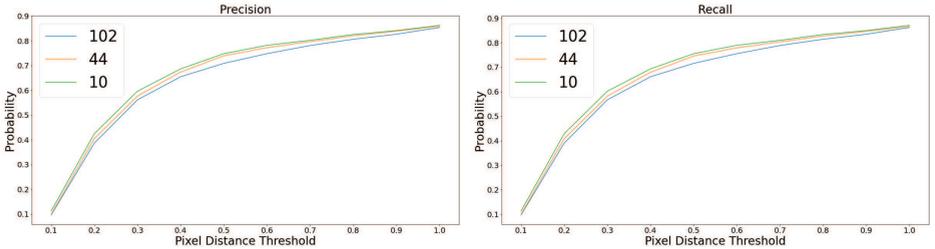}
\caption{Quantitative comparison of plot data extraction with different $\Delta_s$. }
\label{fig: quantitative_comparison_delta_s}
\end{figure}

There are some existing problems with the implementation of the optical flow based algorithm. One problem is that we need to pick a proper value for  $\Delta_s$ in order to have a successful plot line detection. As shown in Fig. \ref{fig: delta_s}, if the $\Delta_s$ is too large (top row), the proposed method fails to inpaint correct misclassified foreground pixels, if the $\Delta_s$ is too small, the proposed method fails to compensate the error from optical flow estimation in case of sudden gradient change (e.g. peak). We also conduct experiments to quantitatively evaluate the performance of the proposed method with different $\Delta_s$. As shown in Fig. \ref{fig: quantitative_comparison_delta_s}, a smaller value of $\Delta_s$ is likely to produce a better detection result.


\section{Appendix B: Ablation study}
\label{sec:appendix_B}

\subsection{Edge-based constraint}

Qualitative comparison on axis alignment with/without the edge-based constraint demonstrates the superiority of the proposed method over the conventional CNN-based object detector. Moreover, we introduce a mean absolute distance between the estimated axes and the real axes to quantitatively measure the axis misalignment. Here we use $(x_{pred}, y_{pred})$ and $(x_{gt}, y_{gt})$ to denote the estimated and ground truth point of origin of the coordinates, respectively. Then the axis misalignment is computed as
\begin{equation}
D^{misalign} = \frac{1}{N}\sum_i|x^i_{pred} - x^i_{gt}| + |y^i_{pred} - y^i_{gt}|
\label{eq: axis_misalignment}
\end{equation}
where $N$ denotes the number of graph images in the testing set. 

We measure the axis misalignment of the detection results using three different anchor-free object detection models, with and without the edge-based constraint. As shown in Table. \ref{tab: axis misalignment}, FCOS \cite{tian2019fcos} outperforms the other detectors, of which the axis misalignment is only 1.49 pixels. The refinement with the edge-based constraint suppresses the axis misalignment, with $\sim$10\% improvement for FCOS and $\sim$47\% improvement for the other detectors.

\begin{table}
\setlength{\belowcaptionskip}{-15pt}
\begin{center}
\begin{tabular}{|l|c|c}
\hline
Method & Refined & Axis Misalignment\\ 
\hline\hline
FCOS \cite{tian2019fcos} & No & 1.49 \\
 & Yes & 1.33 \\
\hline
FreeAnchor \cite{zhang2021learning} & No & 4.65 \\
 & Yes & 2.47 \\
\hline
GuidedAnchor \cite{wang2019region} & No & 3.03 \\
 & Yes & 1.60 \\
\hline
\end{tabular}
\end{center}
\caption{Axis misalignment with different anchor-free object detection models.}
\label{tab: axis misalignment}
\end{table}

\begin{figure}[!t]
\setlength{\belowcaptionskip}{-15pt}
\setlength{\abovecaptionskip}{-20pt}
\centering
\includegraphics[width=0.95\columnwidth]{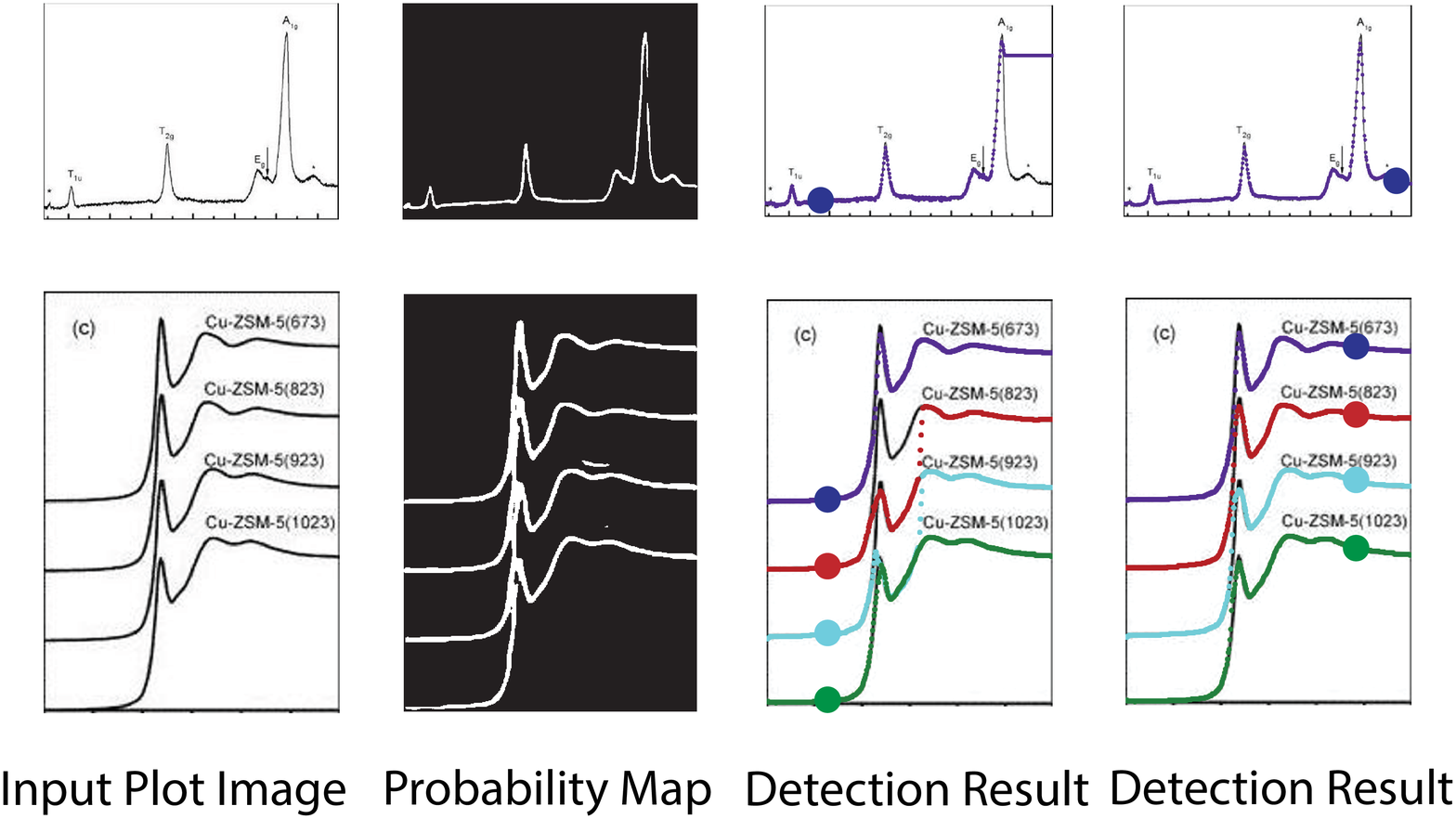}
\caption{Results of plot line detection with different start position. Start positions are highlighted in the detection result images.}
\label{fig: start_posi}
\end{figure}
\subsection{Different start positions for plot line detection}

\begin{figure}[!t]
\setlength{\belowcaptionskip}{-10pt}
\setlength{\abovecaptionskip}{-15pt}
\centering
\includegraphics[width=0.95\columnwidth]{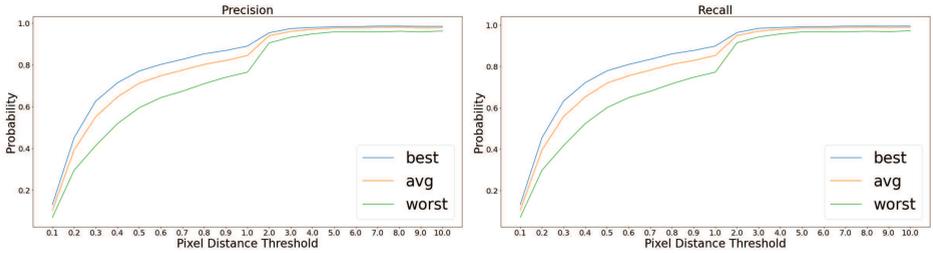}
\caption{Quantitative comparison of plot data extraction with different start positions. }
\label{fig: quantitative_comparison_start_posi}
\end{figure}

A good start position matters in the proposed optical flow method, especially in case that there are sharp peaks in the plot image or misclassified foreground/background pixels in the probability map. Intuitively, we select the start position at places where the gradients are small. As shown in Fig. \ref{fig: start_posi}, the top row shows that misclassified foreground pixels break the continuity of the plot line, which hinders the ability of tracking the motion of the point from one side. In the bottom row, there are sharp peaks in the plot image and significant overlap between different plot lines in the probability map, making it difficult to apply the optical flow method from the left side of the peak. Also, we conduct experiments to quantitatively measure the how the selection of start positions affect the performance. In particular, we randomly select $\sim 20$ start positions in each plot image and apply the proposed method to detect plot lines. We measure the best/average/worst performance of plot line detection with these start positions, as shown in Fig. \ref{fig: quantitative_comparison_start_posi}. Clearly, selecting a proper start position is very important to the success of the algorithm.

\subsection{Different losses for plot line detection}
\begin{figure}[!t]
\setlength{\belowcaptionskip}{-10pt}
\setlength{\abovecaptionskip}{-15pt}
\centering
\includegraphics[width=0.95\columnwidth]{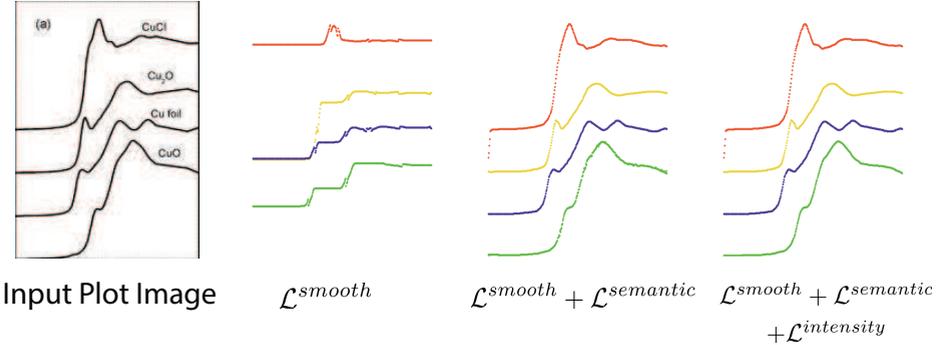}
\caption{Visual comparison of plot data extraction with different losses. }
\label{fig: visual_comparison_diff_loss}
\end{figure}

To have a better understanding over different loss terms in the plot data extraction module, we conduct experiments to determine the difference between the detection results with different loss terms. According to the implementation in Sec. \ref{sec:appendix_A}, the smoothness term produces the optical flow estimation over the next position, but it is likely to be affected by errors in gradients estimation. As shown in Fig. \ref{fig: visual_comparison_diff_loss}, the left most image is the input plot image and the remaining images are the results of the plot line detection with different losses. With the smoothness term only, the detection performance is poor (i.e. middle left image) due to the noise in the plot images which produces inaccurate gradient map.  The semantic loss term is able to compensate for the errors in the optical flow estimation, producing significant improvements (i.e. middle right image ). However, some glitches are still noticeable in the detection result (i.e. the green plot in middle right image). Finally, the intensity constraint term helps refine the detection results (i.e. right most image) by searching for the best intensity match in the neighborhood.

\bibliography{egbib}
\end{document}